**Title: Toward an Automatic System for Computer-Aided Assessment in Facial Palsy**

**Author List:** Diego L. Guarin, PhD[1,2]; Yana Yunusova, PhD, CCC-SLP[1,3,4]; Babak Taati, PhD[1,5,6]; Joseph R Dusseldorp MBBS, MS, FRACS[7]; Suresh Mohan, MD[2]; Joana Tavares, MD[8], Martinus M. van Veen, MD[2]; Emily Fortier[2]; Tessa A. Hadlock, MD[2]; and Nate Jowett, MD[2]

[1] KITE | Toronto Rehabilitation Institute – University Health Network, Toronto, ON, Canada.
[2] Department of Otolaryngology/Head and Neck Surgery, Massachusetts Eye and Ear Infirmary and Harvard Medical School, Boston, USA
[3] Department of Speech Language Pathology, University of Toronto, Toronto, ON, Canada.
[4] Hurvitz Brain Sciences Program, Sunnybrook Research Institute, Ontario, Canada
[5] Department of Computer Science, University of Toronto, Toronto, ON, Canada.
[6] Institute of Biomaterials and Biomedical Engineering, University of Toronto, Toronto, ON, Canada.
[7] Department of Plastic and Reconstructive Surgery, Royal Australasian College of Surgeons and University of Sydney, Sydney, Australia
[8] Faculty of Health Sciences, Brasilia University, Brasilia, Brazil

**Corresponding Author:**

Diego L. Guarin, PhD
Toronto Rehabilitation Institute – University Health Network
550 University Avenue
Toronto, ON M5G 2A2 (Canada)
Diego.Guarin@uhn.ca

**Financial Disclosure Statement:** none

**Abstract word count**: 344/350

**Manuscript word count:** 2582 / 3000

## Key Findings (words = 92/100)

**Question:** Can recent developments in machine learning and computer vision be used to develop an objective and automatic system for computer-aided assessment in facial palsy?

**Findings:** In this research article, we found that by using a relatively small number of manually annotated photographs for a patient specific database it is possible to obtain significant improvement in the accuracy of facial measurements provided by a popular machine learning algorithm.

**Meaning:** The results presented in the article represent the first steps towards the development of an automatic system for computer-aided assessment in facial palsy.

Abstract (words = 344 / 350)

**Importance:** Quantitative assessment of facial function is challenging, and subjective grading scales such as House-Brackmann, Sunnybrook, and eFACE have well-recognized limitations. Machine learning approaches to facial landmark localization carry great clinical potential as they enable high-throughput automated quantification of relevant facial metrics from photographs and videos. However, the translation from research settings to clinical application still requires important improvements.

**Objective:** To develop a novel machine learning algorithm for fast and accurate localization of facial landmarks in photographs of facial palsy patients and utilize this technology as part of an automated computer-aided diagnosis system.

**Design, Setting, and Participants:** Portrait photographs of eight expressions obtained from 200 facial palsy patients and 10 healthy participants were manually annotated, by localizing 68 facial landmarks in each photograph, by three trained clinicians using a custom graphical user interface. A novel machine learning model for automated facial landmark localization was trained using this disease-specific database. Algorithm accuracy was compared to manual markings and the output of a model trained using a larger database consisting only of healthy subjects.

**Main Outcomes and measurements**: Root mean square error normalized by the inter-ocular distance (NRMSE) of facial landmark localization between prediction of machine learning algorithm and manually localized landmarks.

**Results:** Publicly available algorithms for facial landmark localization provide poor localization accuracy when applied to photographs of patients compared to healthy controls (NRMSE, 8.56 ± 2.16 vs. 7.09 ± 2.34, p << 0.01). We found significant improvement in facial landmark localization accuracy for the facial palsy patient population when using a model trained with a relatively small number photographs (1440) of patients compared to a model trained using several thousand more images of healthy faces (NRMSE, 6.03 ± 2.43 vs. 8.56 ± 2.16, p << 0.01).

**Conclusions and relevance:**

Retraining a computer vision facial landmark detection model with fewer than 1600 annotated images of patients significantly improved landmark detection performance in frontal view photographs of this population. The new annotated database and facial landmark localization model represent the first steps towards an automatic system for computer-aided assessment in facial palsy.

**Level of Evidence:** 4

## Introduction

Management of facial neuromuscular disorders is hampered by lack of a universal and objective grading system to characterize disease severity, recovery, and response to therapeutic interventions.[1] Quantifying static facial features and displacements occurring with facial expressions is a promising technique for standardizing assessment in facial palsy, whose reported US incidence exceeds 150,000 cases per year.[2] Several methods exist for measuring facial features and movements. Caliper assessments offer high accuracy yet are tedious and must be performed in person. Computer-based techniques to quantify facial displacements are now widely employed [3-7] Early approaches comprised manual identification of facial landmarks on digital images within specialized software, from which relevant distances and angles could be readily calculated. Though such techniques enabled retrospective assessment of facial function, manual tagging of digital images is resource intensive, error prone, and infeasible for dynamic tracking of facial movements from video. To automate measurement of facial displacements, physical markers placed at specific facial landmarks have been employed, and their location tracked using customized software and hardware.[6] Marker-based tracking is limited as manual marker localization is time consuming, subjective, and requires specialized recording conditions.

Machine learning (ML) -based computer vision algorithms enable rapid and fully automated tracking of facial displacements from digital images and videos recorded under typical conditions with consumer-grade cameras. Such facial landmark detection algorithms are usually trained using databases of manually annotated facial photographs. Once trained, these ML algorithms can predict the position of facial landmarks in a new photograph without human intervention, with high accuracy. [8-13] ML algorithms for facial landmark localization are increasingly being used to study facial palsy,[7, 14-17] Parkinson disease,[18] stroke,[19] amyotrophic lateral sclerosis,[20] and dementia.[21] Owing to their training using predominantly normal subjects, current ML models for facial landmark recognition may be biased against patients, and demonstrate inadequate accuracy when presented with faces of patients with neuromuscular disease impacting facial movements and expression [21]. Herein, we hypothesize that training a ML model for facial landmark localization with facial photographs from a disease-specific clinical database will demonstrate improved tracking accuracy when presented with faces of patients with the condition, in comparison to the one trained using a much larger database of normal subjects.

To evaluate our hypothesis, we introduce the first database of annotated clinical photographs of patients with unilateral facial palsy, employ it to evaluate the bias against patients of a popular facial landmark localization algorithm, and train a ML model for automated facial landmark localization in this population. We further demonstrate the utility of an open-access and user-friendly software, Emotrics, for extracting clinically relevant measurements from photographs and videos in automated fashion using the ML model trained herein.

## Methods

### Automatic Facial Landmark Localization

We employed a popular approach for automatic facial landmark localization in facial photographs known as cascade of regression trees[11, 22-25]. Specifically, we employed the algorithm proposed by Kazemi et al,[11] which provides accurate facial landmark localization results under multitude of pose, illumination, and expression conditions[26] and can process medical images in just a few hundred milliseconds without the use of Graphical Processing Units (GPU) or other specialized hardware.[10]

Implementation of this algorithm for facial landmark localization is readily available in open source machine learning libraries such as OpenCV[27], and Dlib[28]. These implementations were trained using the 300-W dataset from the Intelligent Behavior Understanding Group (iBUG)[12]. This dataset is a concatenation of several freely-available databases (LFPW[29], HELEN[30], AFW[31], and iBUG[32]), and comprises 11500 in-the-wild photographs of over 4000 healthy subjects. Model training has been performed by manually annotating a set of facial landmarks in each photograph using the 68-point Carnegie Mellon University multiple pose, illumination, and expression (Multi-PIE) database approach[33,34]. Manually annotated landmarks outlined the superior border of the brow, the free margin of the upper and lower eyelids, the nasal midline, the nasal base, the mucosal edge and vermillion-cutaneous junction of the upper and lower lips, and the lower two-thirds of the face.

### MEEI Database

We obtained Institutional review board approval to access the Massachusetts Eye and Ear (MEE) Facial Nerve Data Repository, a digital collection of facial photographs and video clips of patients with unilateral facial palsy at the MEE Facial Nerve Center. The patients had consented to use of facial photographs for research purposes. A clinical photographer captured high-resolution photographs (1080 x 720 pixels) of a standardized series of facial expressions used to evaluate facial mimetic function.[35] Figure 1 exemplifies the type of photographs taken from each patient; all images were taken from the frontal view using a digital camera with optimal lighting.

In addition to the patient's photographs, we obtained high-resolution photographs of healthy controls from the MEEI Facial Palsy Photo and Video Standard Set[14] performing the same standardized facial expressions as the patients.

### Photograph Annotation

Three trained clinicians independently annotated the 68 Multi-PIE landmark points for each photograph in the clinical database using a custom graphical user interface[10] (Emotrics software, Mass Eye and Ear, Boston MA). Images were uploaded into Emotrics, which provided an initial estimation of the 68 landmarks points. Landmark positions for each photograph were verified and manually re-positioned as necessary by each marker. Landmark positions of three manual annotators were averaged for each photograph to define ground-truth locations.

### Evaluating Model Bias and Training a New Model for Landmark Localization

Marked photographs were clustered by patient and randomly divided into three non-overlapping groups, all images of the same patients were only in one group. The groups were: - model training (N=180 subjects; equating to 90% of the database or 1440 photographs), validation (N=10 subjects; equating to 5% of the database or 80 photographs, and test (N=10 subjects; equating to 5% of the database or 80 photographs). Healthy control photographs (N=10 subjects, equating to 80 photographs) were used for model testing only. Computer operations were performed in the Python programming language (version 3.6.7) on a Lenovo Thinkpad personal computer (T470, Intel Core i7-7600U processor running at 2.8GHz with 32GB of RAM) that did not include specialized hardware for training and evaluation of machine learning models (GPU or TPU).

### Evaluation of model bias against patients

Evaluation of model bias against patients was performed using publicly available implementation of the ensemble of regression trees algorithm for facial landmark localization proposed by Kazemi et al.[11] Available algorithm was used to estimate the position of 68 facial landmarks in the test group of our database and the photographs of healthy controls; predicted landmarks position were compared to the ground-truth locations provided by the manual annotation procedure. Errors in landmark localization were quantified using the root-mean-square error (RMSE) between ground-truth and algorithm-predicted landmark positions normalized by the inter-ocular distance (NRMSE) [38].

### Training a specialized model for facial landmark localization in patients

Next, photographs from the training group were employed to re-train the ensemble of regression trees model. Photographs from the validation group were used to determine the model parameters that provided the minimum landmark localization error using cross-validation. Standard model parameters such as number of estimators, tree deep, minimum number of samples per leaf, and number of features were selected using a grid search; a total of 400 permutations were assessed. The parameter set yielding the lowest NRMSE for the validation dataset was chosen for model testing and comparison.

The re-trained model was compared against the implementation trained with the 300-W database by computing the NRMSE between the ground-truth and model predicted landmark positions yielded by both models for the test group (patients and controls) of our database.

### Statistical Analysis

Differences between the groups (i.e, healthy vs. patients) were sought using the Kruskal-Wallis one-way analysis of variance; differences between models (i.e. publicly available vs. re-trained using patients' photographs) were sought using the Wilcoxon signed rank test. Statistical significance was considered at $p < 0.01$.

# Results

## MEE Database

All patients that visited the Facial Nerve Center at MEE between October 2017 and October 2018 and provided written consent to use their photographs for research purposes were considered candidates for the database. Patients who demonstrated gross facial deformities or whose facial features were not clearly visible (due to obstruction by clothing or hair) were excluded. Media from 200 adult and pediatric patients with varying facial palsy severity and etiology, totaling 1600 photographs, were collected, manually annotated and employed for algorithm training and testing; Table 1 presents relevant demographic and clinical information.

## Model bias against patients

Figure 2A shows a box and whiskers plow representing the NRMSE yielded by the standard implementation of the ensemble of regression trees model trained with the 300-W database, when applied to photographs of healthy controls and patients with facial palsy. Our results demonstrated that this implementation of the algorithm yielded significantly worst facial landmark localization accuracy, as quantified by the NRMSE, for patients than for healthy subjects ($7.09 \pm 2.34$ vs. $8.56 \pm 2.16$, $p \ll 0.01$).

## Specialized model Accuracy

**Error! Reference source not found.** shows a box and whiskers plot representing the NRMSE yielded by the ensemble of regression trees model re-trained with the MEEI database, when applied to photographs of healthy controls and patients suffering from facial palsy. Our analysis demonstrated that there was no significant difference in the facial landmark localization accuracy between patients and healthy subjects ($6.87 \pm 2.28$ vs. $6.03 \pm 2.43$, $p = 0.03$).

There was no significant difference in the landmark localization accuracy provided by the models trained with the 300-W and MEEI databases when applied to healthy subjects (7.09 ± 2.34 vs. 6.87 ± 2.28, p = 0.162). In contrast, the re-trained model provided significantly improved accuracy when applied to patients (8.56 ± 2.16 vs. 6.03 ± 2.43, p<<0.01). Figure 3 illustrates the improved accuracy of the model trained with the MEEI database over the model trained with the 300-W database. Figure 3 A, C demonstrate obvious localization errors for points defining the facial contour and lips yielded by the 300-W model output; Figure 3 B, D demonstrate than fewer and smaller errors are present for the output of the MEEI model.

## Discussion

Objective quantification of disease severity facilitates improved understanding of disease progression, recovery, and treatment response, and patient-clinician and inter-clinician communication. Facial palsy severity is typically assessed using subjective clinician facial grading systems including the House-Brackmann,[36] Sunnybrook,[37] and eFACE[38] scales. Such approaches are limited by high inter-rater variability, and require considerable training for proper use and interpretation.[39,40] Though more objective methods for quantifying facial displacements have been described, no single tool has achieved widespread use.

In this study we demonstrated that machine learning approaches can provide objective, automatic, and accurate facial measurements in photographs of patients suffering from facial palsy, so that these methods have the potential of disrupting the current clinical practice for diagnosis and assessment of the condition. However, our results demonstrated that publicly available models, trained with databases of healthy subjects, provide significantly worst landmark localization accuracy when applied to photographs of patients. We also demonstrated that by re-training the facial landmark localization model using a small number of photographs from a disease-specific clinical database, it is possible to significantly improve the facial landmark localization accuracy in this patient population. The standardized recording conditions (pose, illumination, expression, and background) of photographs in the clinical database likely explains the observed high accuracy of the model, despite use of a relatively small training dataset. These results supported our hypothesis that a novel model for facial landmark localization trained using disease-specific photographs would demonstrate improved tracking accuracy when presented with faces of patients with the condition, in comparison to one trained using a much larger database of normal subjects.

Finally, we found no significant difference in the landmark localization accuracy yielded by the models trained with the 300-W and MEEI databases when applied to photographs of healthy subjects. This is an unexpected but welcomed result, as it indicates that the new model can be used to track the recovery of a patient across a continuum from highly impaired to normal.

Clinical application - Computer-Aided Assessment in Facial Palsy

The facial landmark localization model trained with the MEEI database was packaged into the latest version of a previously-characterized open-source customized software platform for automatic estimation of clinically-relevant facial metrics – Emotrics.[10] This tool uses landmark positions provided by a designated 68-point facial landmark localization ML model to estimate facial metrics relevant to the field of facial palsy in high-throughput fashion.[10, 14-17] Figure 4 illustrates various facial metrics computed automatically using Emotrics from clinical photographs of three subjects during full-effort smile. Photographs were taken from the MEEI Facial Palsy Photo and Video Standard Set[14] and were not used to train or validate the models presented here. The first subject demonstrates normal facial function and near-symmetric facial metrics. The second subject developed left-sided flaccid facial paralysis following Bell's palsy. The third subject developed left-sided post-paralytic facial palsy and synkinesis following Bell's palsy. Marked differences in oral commissure position and palpebral fissure heights between sides and subjects are readily quantified.

## Limitations

There are several limitations with the facial landmark localization model described herein. The database comprises patient photographs from a single center and reflects its demographics. The database includes more female (N=135) than male (N=65) patients, and its racial demographic is mostly white (N=160), with small representation of minorities, including Hispanic (N=15), Asian (N=14), and black (N=11). Additionally, the database comprises more middle-aged adults (age group = (24, 64] years, N=142) than younger adults (age group = (18, 25] years, N=12), older adults (age group = 64+ years, N=41), and children (age group = (0, 18] years, N=5). Other sources of model bias such as the presence of facial hair were not assessed. Non-symmetric distribution of patient demographics might lead to prediction error bias; for example, the model might demonstrate higher performance among adult middle-aged white women as they comprised the largest cohort of the training dataset. The model is further limited in that recording conditions (pose, illumination, and expression) were specific to our clinical center. While patient pose (requiring frontal view of face with neutral roll, tilt, and yaw) and expression may be readily standardized across clinical centers, illumination conditions are more challenging to standardize and their impact on model accuracy has yet to be assessed. Further work will seek to expand the training dataset to include patient photographs from multiple clinical centers to improve model accuracy across a wider range of patient demographics and disease severities. The facial landmark localization model was applied only to patient photographs of eight fixed facial expressions. Future work will seek to assess the performance of this model for dynamic tracking of facial landmarks during expression from databased videos of patients with unilateral facial palsy.

### Availability

Emotrics and the two ML models for facial landmark localization discussed here are freely available online on GitHub (www.github.com/dguari1) and the Sir Charles Bell Society website (www.sircharlesbell.com). Emotrics software is open-access and open-source; the Python-based code can be modified as desired. Request from research institutes to share the data will be reviewed on case-by-case basis by the MEEI Institutional Review Board. Due to patient privacy concerns, the annotated database cannot be shared online.

## Conclusions

We introduced the first manually annotated database of standardized pose, illumination, and expression photographs among patients with unilateral facial palsy. Using this dataset, we demonstrated that a ML model for automatic facial landmark localization in this patient population outperforms a model trained using a much larger dataset of healthy subjects. We demonstrated the clinical utility of this approach in the quantification of facial palsy disease severity from databased photographs and characterized an open-access software tool that facilitates rapid calculation of relevant facial metrics in this patient population.

# References

1. Hadlock, T.A. and L.S. Urban, Toward a universal, automated facial measurement tool in facial reanimation. Arch Facial Plast Surg, 2012;14(4):277-82.
2. Bleicher, J.N., S. Hamiel, J.S. Gengler, and J. Antimarino, A survey of facial paralysis: etiology and incidence. Ear Nose Throat J, 1996;75(6):355-8.
3. Bray, D., D.K. Henstrom, M.L. Cheney, and T.A. Hadlock, Assessing outcomes in facial reanimation: evaluation and validation of the SMILE system for measuring lip excursion during smiling. Arch Facial Plast Surg, 2010;12(5):352-4.
4. Coulson, S.E., G.R. Croxson, and W.L. Gilleard, Three-dimensional quantification of "still" points during normal facial movement. Ann Otol Rhinol Laryngol, 1999;108(3):265-8.
5. Frey, M., M. Michaelidou, C.H. Tzou, I. Pona, M. Mittlbock, H. Gerber, and E. Stussi, Three-dimensional video analysis of the paralyzed face reanimated by cross-face nerve grafting and free gracilis muscle transplantation: quantification of the functional outcome. Plast Reconstr Surg, 2008;122(6):1709-22.
6. Gerós, A., R. Horta, and P. Aguiar, Facegram–Objective quantitative analysis in facial reconstructive surgery. Journal of biomedical informatics, 2016;61:1-9.
7. Guo, Z., G. Dan, J. Xiang, J. Wang, W. Yang, H. Ding, O. Deussen, and Y. Zhou, An unobtrusive computerized assessment framework for unilateral peripheral facial paralysis. IEEE journal of biomedical and health informatics, 2018;22(3):835-841.
8. Cootes, T.F., G.J. Edwards, and C.J. Taylor, Active appearance models. IEEE Transactions on pattern analysis and machine intelligence, 2001;23(6):681-685.
9. Fan, H. and E. Zhou, Approaching human level facial landmark localization by deep learning. Image and Vision Computing, 2016;47:27-35.
10. Guarin, D.L., J.R. Dusseldorp, H. A., and N. Jowett, Automated Facial Measurements in Facial Paralysis: A Machine Learning Approach. JAMA Facial Plastic Surgery, 2017;20(4):335-337.
11. Kazemi, V. and J. Sullivan. One millisecond face alignment with an ensemble of regression trees. in Proceedings of the IEEE Conference on Computer Vision and Pattern Recognition. 2014.
12. Sagonas, C., G. Tzimiropoulos, S. Zafeiriou, and M. Pantic. 300 faces in-the-wild challenge: The first facial landmark localization challenge. in Proceedings of the IEEE International Conference on Computer Vision Workshops. 2013.
13. Zhang, Z., P. Luo, C.C. Loy, and X. Tang. Facial landmark detection by deep multi-task learning. in European Conference on Computer Vision. 2014. Springer.
14. Greene, J.J., J. Tavares, D.L. Guarin, E. Fortier, M. Robinson, J.R. Dusseldorp, O. Quatela, N. Jowett, and T. Hadlock The Spectrum of Facial Palsy: The MEEI Facial Palsy Photo & Video Standard Set [published online April 25, 2019]. .The Laryngoscope. DOI: 10.1002/lary.27986.

15. Greene, J.J., J. Tavares, D.L. Guarin, N. Jowett, and T. Hadlock, Surgical Refinement Following Free Gracilis Transfer for Smile Reanimation. Annals of plastic surgery, 2018;81(3):329-334.
16. Greene, J.J., J. Tavares, S. Mohan, N. Jowett, and T. Hadlock, Long-Term Outcomes of Free Gracilis Muscle Transfer for Smile Reanimation in Children. The Journal of pediatrics, 2018;202:279-284. e2.
17. Jowett, N. and T.A. Hadlock, Facial Palsy: Diagnostic and Therapeutic Management. Otolaryngologic Clinics of North America, 2018;51(6):xvii-xviii.
18. Li, M.H., T.A. Mestre, S.H. Fox, and B. Taati, Vision-based assessment of parkinsonism and levodopa-induced dyskinesia with pose estimation. Journal of NeuroEngineering and Rehabilitation, 2018;15(1):97.
19. Lanz, C., B.S. Olgay, J. Denzler, and H.-M. Gross. Facial Landmark Localization and Feature Extraction for Therapeutic Face Exercise Classification. in International Conference on Computer Vision, Imaging and Computer Graphics. 2013. Springer.
20. Bandini, A., J.R. Green, B. Taati, S. Orlandi, L. Zinman, and Y. Yunusova. Automatic detection of amyotrophic lateral sclerosis (ALS) from video-based analysis of facial movements: speech and non-speech tasks. in 2018 13th IEEE International Conference on Automatic Face & Gesture Recognition (FG 2018). 2018. IEEE.
21. Taati, B., S. Zhao, A.B. Ashraf, A. Asgarian, M.E. Browne, K.M. Prkachin, A. Mihailidis, and T. Hadjistavropoulos, Algorithmic Bias in Clinical Populations – Evaluating and Improving Facial Analysis Technology in Older Adults with Dementia. IEEE Access, 2019:1-1.
22. Dollár, P., P. Welinder, and P. Perona. Cascaded pose regression. in 2010 IEEE Computer Society Conference on Computer Vision and Pattern Recognition. 2010. IEEE.
23. Johnston, B. and P. de Chazal, A review of image-based automatic facial landmark identification techniques. EURASIP Journal on Image and Video Processing, 2018;2018(1):86.
24. Sun, Y., X. Wang, and X. Tang. Deep convolutional network cascade for facial point detection. in Proceedings of the IEEE conference on computer vision and pattern recognition. 2013.
25. Tzimiropoulos, G. Project-out cascaded regression with an application to face alignment. in Proceedings of the IEEE Conference on Computer Vision and Pattern Recognition. 2015.
26. Bulat, A. and G. Tzimiropoulos. How far are we from solving the 2d & 3d face alignment problem?(and a dataset of 230,000 3d facial landmarks). in Proceedings of the IEEE International Conference on Computer Vision. 2017.
27. Baksheev, A., V. Erumihov, K. Kornyakov, and K. Pulli, *Realtime computer vision with opencv*. 2012, Queue.
28. King, D.E., Dlib-ml: A machine learning toolkit. Journal of Machine Learning Research, 2009;10(Jul):1755-1758.
29. Belhumeur, P.N., D.W. Jacobs, D.J. Kriegman, and N. Kumar, Localizing parts of faces using a consensus of exemplars. IEEE transactions on pattern analysis and machine intelligence, 2013;35(12):2930-2940.

30. Le, V., J. Brandt, Z. Lin, L. Bourdev, and T.S. Huang. Interactive facial feature localization. in European conference on computer vision. 2012. Springer.
31. Ramanan, D. and X. Zhu. Face detection, pose estimation, and landmark localization in the wild. in Proceedings of the 2012 IEEE Conference on Computer Vision and Pattern Recognition (CVPR). 2012. Citeseer.
32. Sagonas, C., G. Tzimiropoulos, S. Zafeiriou, and M. Pantic. A semi-automatic methodology for facial landmark annotation. in Proceedings of the IEEE conference on computer vision and pattern recognition workshops. 2013.
33. Gross, R., I. Matthews, J. Cohn, T. Kanade, and S. Baker, Multi-pie. Image and Vision Computing, 2010;28(5):807-813.
34. Sim, T., S. Baker, and M. Bsat. The CMU pose, illumination, and expression (PIE) database. in Proceedings of Fifth IEEE International Conference on Automatic Face Gesture Recognition. 2002. IEEE.
35. Santosa, K.B., A. Fattah, J. Gavilán, T.A. Hadlock, and A.K. Snyder-Warwick, Photographic standards for patients with facial palsy and recommendations by members of the Sir Charles Bell Society. JAMA facial plastic surgery, 2017;19(4):275-281.
36. House, W., Facial nerve grading system. Otolaryngol Head Neck Surg, 1985;93:184-193.
37. Ross, B.G., G. Fradet, and J.M. Nedzelski, Development of a sensitive clinical facial grading system. Otolaryngol Head Neck Surg, 1996;114(3):380-6.
38. Banks, C.A., P.K. Bhama, J. Park, C.R. Hadlock, and T.A. Hadlock, Clinician-graded electronic facial paralysis assessment: the eFACE. Plastic and reconstructive surgery, 2015;136(2):223e-230e.
39. Banks, C.A., N. Jowett, B. Azizzadeh, C. Beurskens, P. Bhama, G. Borschel, C. Coombs, S. Coulson, G. Croxon, and J. Diels, Worldwide testing of the eFACE facial nerve clinician-graded scale. Plastic and reconstructive surgery, 2017;139(2):491e-498e.
40. Reitzen, S.D., J.S. Babb, and A.K. Lalwani, Significance and reliability of the House-Brackmann grading system for regional facial nerve function. Otolaryngology—Head and Neck Surgery, 2009;140(2):154-158.

# Figures

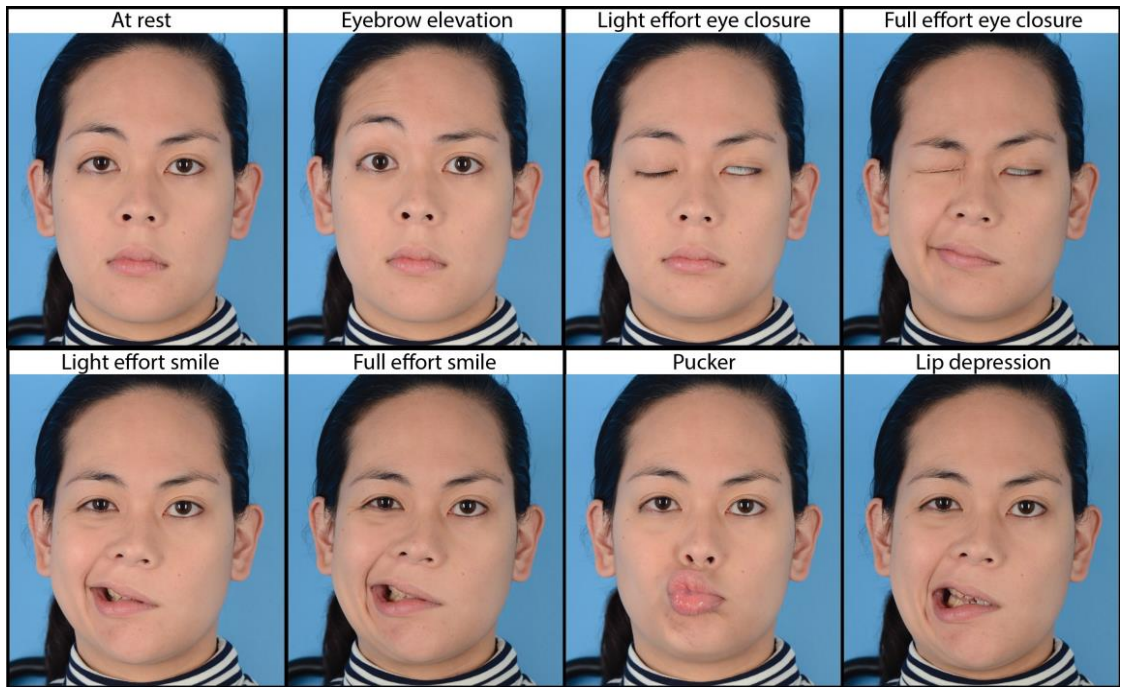

*Figure 1. Example of the standardized set of facial expressions among facial palsy patients in the annotated database. These facial expressions enable global and zonal evaluation of facial function from still photographs.*

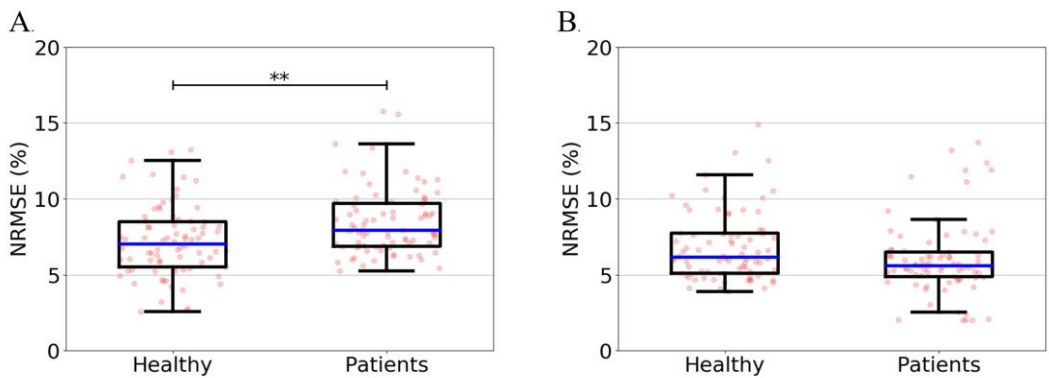

*Figure 2. Normalized Root Mean Square Error of facial landmark localization A. Model trained with photographs of healthy subjects applied to photographs of healthy subjects and patients. B. Model trained with photographs of patients suffering from facial palsy applied to photographs of healthy subjects and patients.*

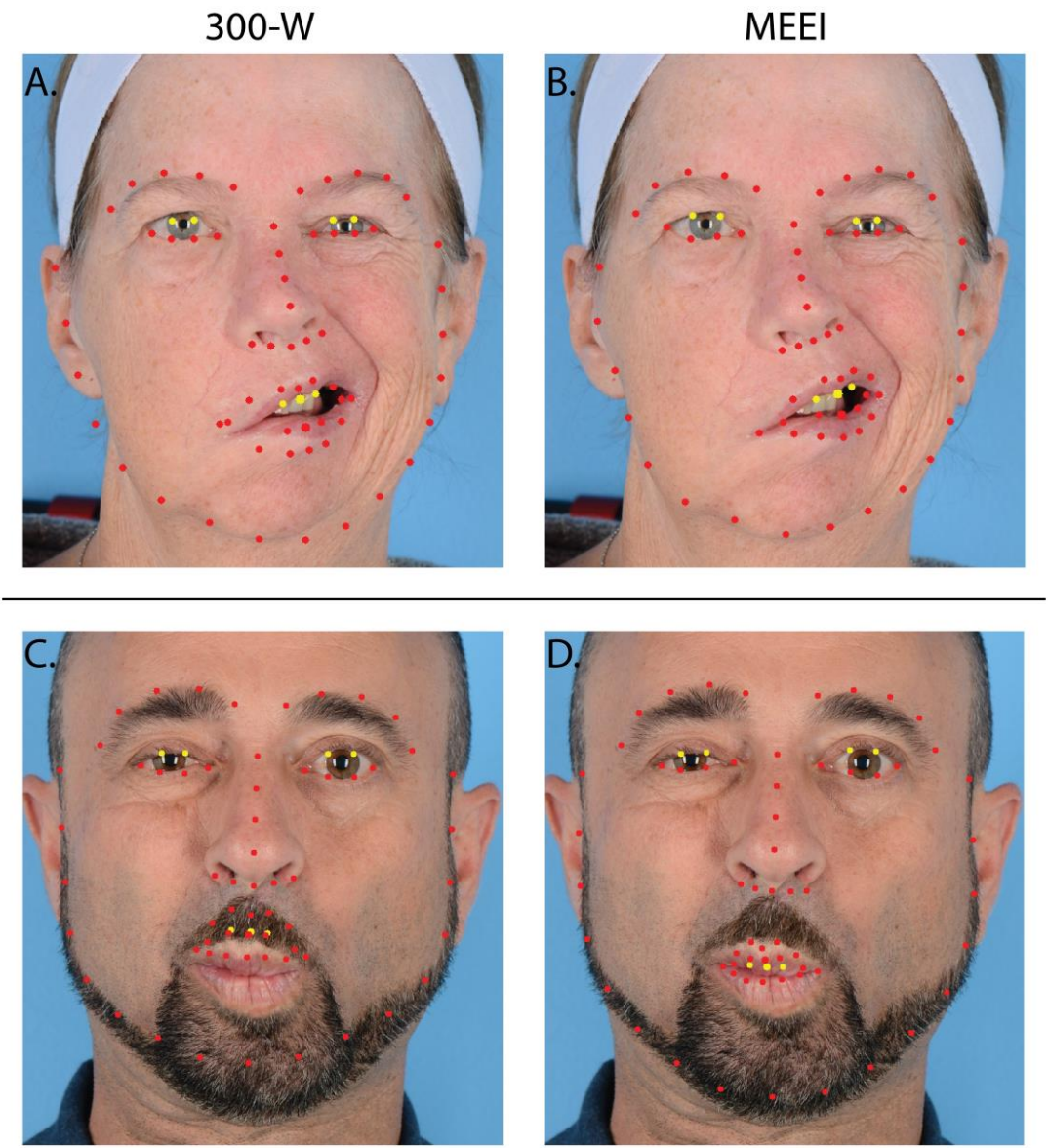

*Figure 3. Comparison of automatic facial landmark localizations by models trained using the 300-W (A and C) and MEEI (B and D) databases among two patients with facial palsy in the validation group.*

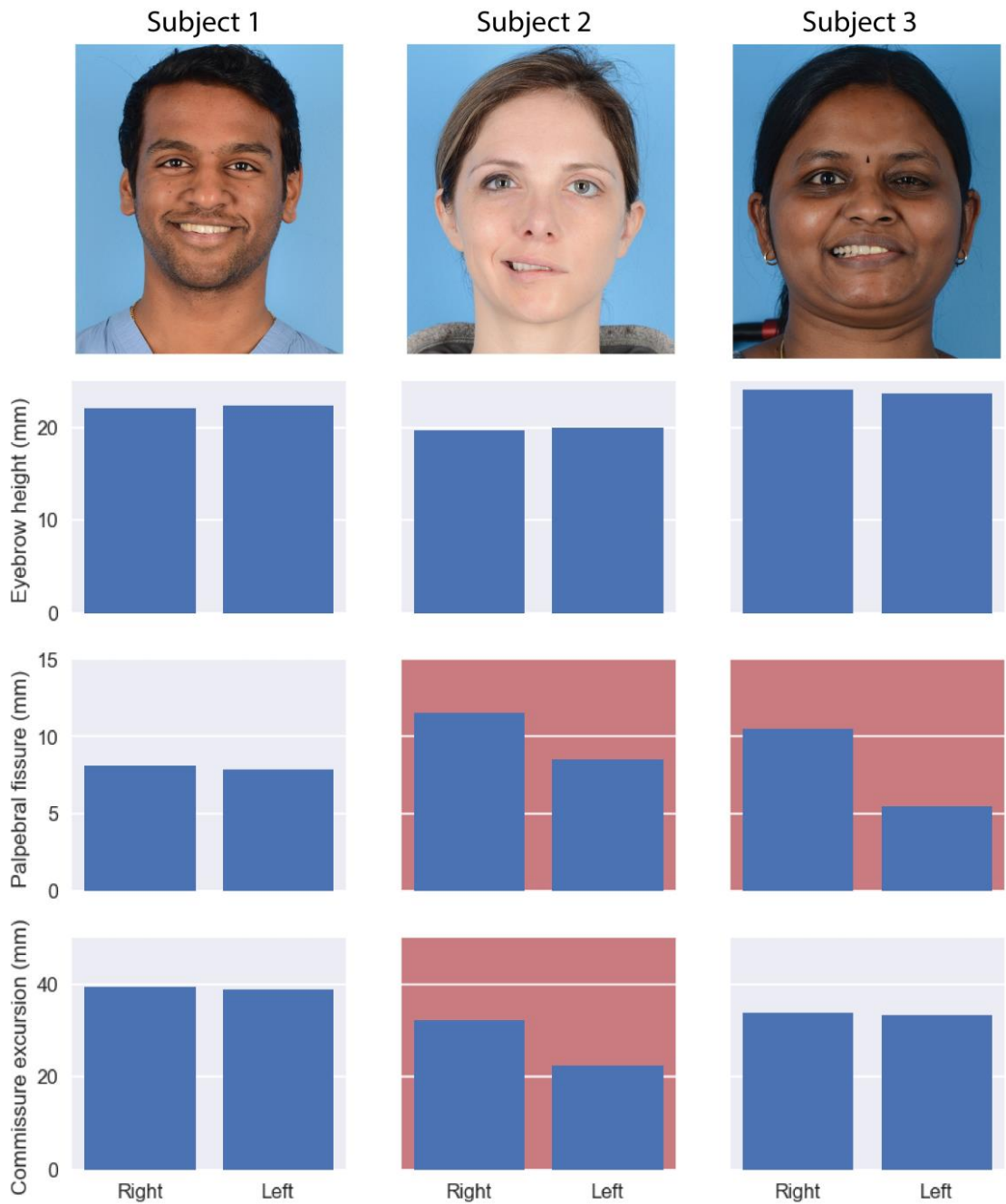

*Figure 4. Examples of automated facial metrics calculated using Emotrics software. Measurements were computed using estimated landmark position provided by the ML model trained using photographs of facial palsy patients. Red background indicates facial measurements that are markedly different among facial sides. Subject 1 is normal, Subject 2 suffers from flaccid facial palsy, and Subject 3 suffers from post-paralytic facial palsy. Presented measures include eyebrow height (vertical distance from the mid-pupillary point to the superior border of the brow), palpebral fissure height (vertical distance between central portions of upper and lower lid margins), and commissure excursion (distance from the facial midline at the lower lip vermilion junction to the oral commissure).*

# Tables

## Table 1. Patients' Demographics

| Demographics | | N |
|---|---|---|
| **Age** | | |
| | range (years) | 7-89 |
| | mean ± std (years) | 48.9 ± 17.1 |
| **Gender** | | |
| | Female | 135 |
| | Male | 65 |
| **Race** | | |
| | White | 160 |
| | Hispanic | 15 |
| | Asian | 14 |
| | Black | 11 |
| **Etiology** | | |
| Infectious | Bell's palsy | 65 |
| | Ramsay Hunt syndrome | 14 |
| | Lyme disease | 12 |
| | Pregnancy-associated Bell's palsy | 7 |
| | Zoster sine herpete | 6 |
| | Meningitis | 1 |
| Congenital | Congenital facial palsy | 5 |
| Otologic | Cholesteatoma | 2 |
| Neoplastic | Vestibular schwannoma | 21 |
| | Parotid neoplasm | 14 |
| | Facial nerve schwannoma | 8 |
| | Brainstem neoplasm | 4 |
| | CNS metastasis | 1 |
| | Basal cell carcinoma | 1 |
| | Trigeminal nerve neoplasm | 1 |
| Trauma | Trauma | 6 |
| Iatrogenic | Iatrogenic | 7 |
| Neurologic | Hemifacial spasm | 1 |
| | Multiple sclerosis | 1 |
| Vascular | Brainstem Stroke | 1 |
| | Venous malformation | 1 |
| | Cavernous brainstem hemangioma | 1 |